\documentclass[journal,twoside,web]{IEEEtran}
\usepackage{cite}
\usepackage{algorithm}
\usepackage{algorithmic}
\usepackage[caption=false,font=normalsize,labelfont=sf,textfont=sf]{subfig}
\usepackage{amsmath,amssymb,amsfonts}
\usepackage{graphicx}
\usepackage{float}
\usepackage{stfloats}
\usepackage{textcomp}
\usepackage{multirow}
\usepackage{tabularx}
\usepackage{hyperref}
\usepackage{url}
\usepackage{tikz}
\usepackage{xcolor}
\usepackage{colortbl}
\usepackage{comment}
\usepackage{booktabs}
\usepackage{array}
\usepackage{verbatim}

\def\BibTeX{{\rm B\kern-.05em{\sc i\kern-.025em b}\kern-.08em
    T\kern-.1667em\lower.7ex\hbox{E}\kern-.125emX}}
\usepackage{balance}

\begin{document}

\title{Mamba-Based Modality Disentanglement Network for Multi-Contrast MRI Reconstruction}
\author{Weiyi Lyu, Xinming Fang, Jun Wang, Jun Shi, Guixu Zhang, Juncheng Li* 
\thanks{Corresponding author: Juncheng Li (jcli@cs.ecnu.edu.cn)}
\thanks{Weiyi Lyu, Xinming Fang, Jun Wang, and Jun Shi are with the School of Communication and Information Engineering, Shanghai University, Shanghai 200444, China.}
\thanks{Guixu Zhang and Juncheng Li are with the School of Computer Science and Technology, East China Normal University, Shanghai 200062, China.}}

\maketitle

\begin{abstract}
Magnetic resonance imaging (MRI) is a cornerstone of modern clinical diagnosis, offering unparalleled soft-tissue contrast without ionizing radiation. However, prolonged scan times remain a major barrier to patient throughput and comfort. Existing accelerated MRI techniques often struggle with two key challenges: (1) failure to effectively utilize inherent K-space prior information, leading to persistent aliasing artifacts from zero-filled inputs; and (2) contamination of target reconstruction quality by irrelevant information when employing multi-contrast fusion strategies. To overcome these challenges, we present MambaMDN, a dual-domain framework for multi-contrast MRI reconstruction. Our approach first employs fully-sampled reference K-space data to complete the undersampled target data, generating structurally aligned but modality-mixed inputs. Subsequently, we develop a Mamba-based modality disentanglement network to extract and remove reference-specific features from the mixed representation. Furthermore, we introduce an iterative refinement mechanism to progressively enhance reconstruction accuracy through repeated feature purification. Extensive experiments demonstrate that MambaMDN can significantly outperform existing multi-contrast reconstruction methods.
\end{abstract}


\section{Introduction}
\IEEEPARstart{M}{agnetic} resonance imaging (MRI) is a vital diagnostic tool, providing unparalleled soft-tissue contrast without ionizing radiation—making it indispensable for neurological, musculoskeletal, and oncological imaging. Unlike Computed Tomography (CT), X-ray, or Positron Emission Tomography (PET), MRI avoids radiation exposure, enabling safer repeated scans for chronic disease monitoring and pediatric cases. However, its widespread clinical utility is hampered by three fundamental limitations, including 1) K-space sampling constraints, 2) multi-sequence protocols, and 3) physically limited timing. These factors contribute to patient discomfort, motion artifacts, and reduced clinical throughput, particularly serious in resource-limited healthcare settings.


 

\begin{figure}[!t]
\centering
\subfloat[Multi-contrast MRI reconstruction using fusion strategy.\label{single-recon}]{
    \includegraphics[width=1\linewidth]{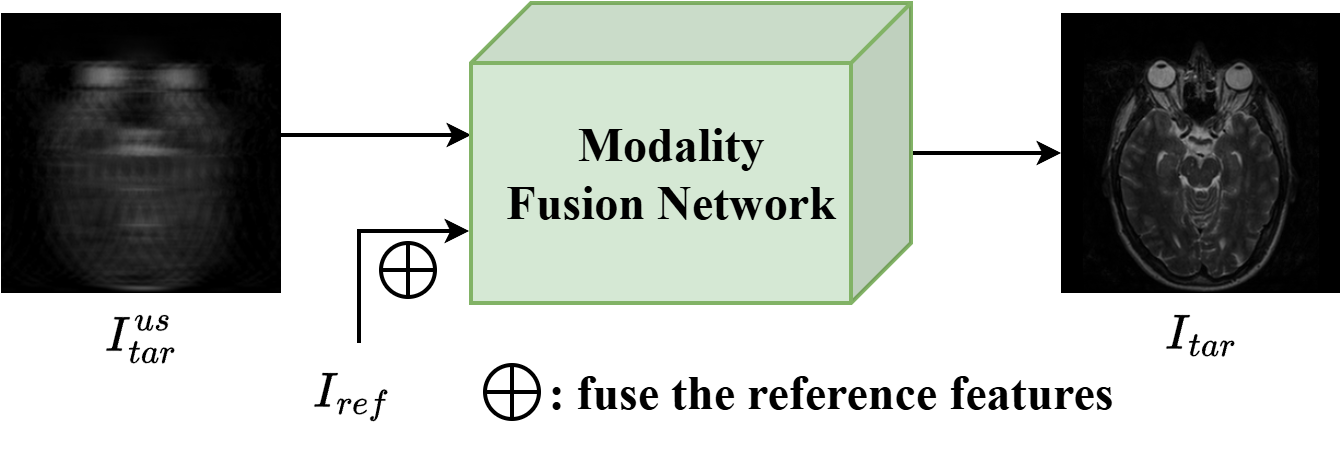}
}\hfill
\subfloat[Our proposed modality disentanglement strategy.\label{multi-recon}]{
    \includegraphics[width=1\linewidth]{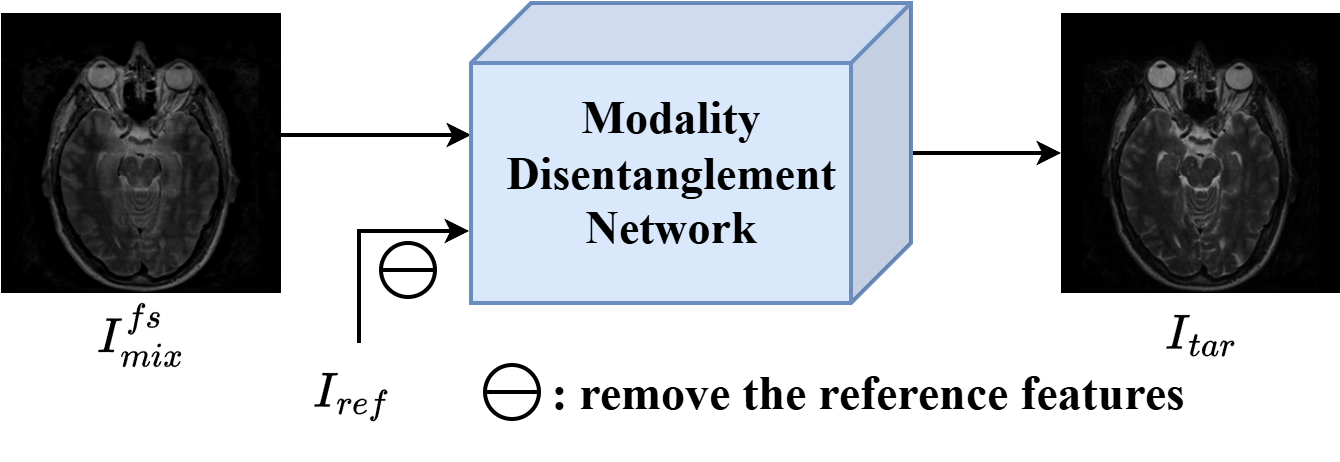}
}
\caption{Comparison of multi-contrast MRI reconstruction strategies. (a) Fusion-based approaches fuse the undersampled target image $I_{tar}^{us}$ with the high-quality reference image $I_{ref}$, which may introduce irrelevant features. (b) Our proposed disentanglement-based method first complements the undersampled target K-space using the reference modality, generating a structure-preserved but modality-mixed image $I_{mix}^{fs}$. A modality disentanglement network then removes reference-specific features to yield a clean target reconstruction $I_{tar}$.}
\label{fusionVSdisentanglement}
\end{figure}

Fast MRI reconstruction techniques address a critical clinical need by enabling high-quality image recovery from undersampled k-space data, significantly reducing acquisition times while maintaining diagnostic utility. Traditional approaches, such as Parallel Imaging (PI)~\cite{sodickson1997simultaneous,pruessmann1999sense} and Compressed Sensing (CS)~\cite{donoho2006compressed}, leverage the sparsity of transformed domains and incorporate prior information into regularized optimization frameworks. These methods also exploit the redundancy among multiple coils to reconstruct the full-resolution image. However, despite their theoretical foundations, traditional algorithms often suffer from compromised image fidelity and long reconstruction times, hindering their practical deployment in real-world clinical environments.

Modern deep learning approaches~\cite{lecun2015deep} have improved reconstruction fidelity and computational efficiency, making them increasingly viable for clinical adoption. Some deep learning frameworks~\cite{wang2016accelerating,qin2018convolutional,quan2018compressed,feng2021task,huang2022swin} treat MRI reconstruction as an image-domain task, neglecting the complex-valued nature of raw k-space data and its vital phase information, which is indispensable for functional and quantitative MRI applications. Therefore, dual-domain methods~\cite{eo2018kiki,souza2019hybrid,wang2022dimension,wang2024ddc} attempt to bridge this gap by jointly optimizing image and frequency-domain representations, yet challenges persist in utilizing different domain information. Overcoming these limitations is essential for translating accelerated MRI into routine practice, particularly for time-sensitive diagnoses (e.g., stroke or dynamic contrast-enhanced studies) and resource-constrained settings.

Another promising direction is multi-contrast MRI reconstruction, which leverages complementary information across different contrast modalities. Since some modalities (e.g., T1) can be acquired faster, they may serve as references to assist the reconstruction of slower modalities (e.g., T2). As shown in Fig.~\ref{fusionVSdisentanglement} (a), multi-contrast reconstruction approaches typically employ a dual-branch architecture that processes undersampled target images alongside fully-sampled reference images, subsequently fusing their complementary information. While existing methods primarily concentrate on optimizing fusion strategies~\cite{lyu2025fast}, this paradigm inherently introduces extraneous information that may compromise reconstruction fidelity. In contrast, our approach fundamentally rethinks this framework by directly utilizing the reference modality to complete the undersampled target K-space data, thereby transitioning from feature fusion to feature disentanglement as shown in Fig.~\ref{fusionVSdisentanglement} (b). This paradigm shift guarantees structurally preserved inputs with suppressed artifacts for subsequent reconstruction stages.

\begin{figure}[t]
\centering
\includegraphics[width=0.48\textwidth]{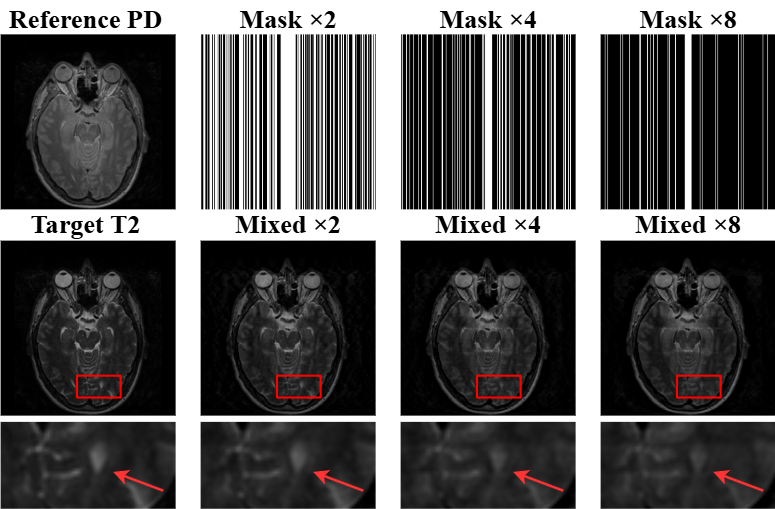} 
\caption{Visualization of modality-mixed images generated under different accelerations. As the undersampling factor increases, more information is borrowed from the reference modality, resulting in a hybrid reconstruction that exhibits enhanced PD-like contrast (e.g., higher gray matter intensity and reduced T2-specific fluid signal).} 
\label{KCMresult} 
\end{figure}

Specifically, we propose \textbf{MambaMDN}, a new dual-domain framework for multi-contrast MRI reconstruction that integrates the K-space complementation strategy with an image-domain modality disentanglement module. In the K-space domain, we explicitly leverage the fully sampled reference modality to fill the missing K-space lines in the undersampled target measurement. This step is conceptually related to the Keyhole method\cite{van1993keyhole} used in dynamic MRI, where central K-space from a reference frame of the same modality (e.g., before or after contrast agent injection) is substituted to accelerate acquisition. However, unlike Keyhole, which assumes the reference and target share the same contrast and only differ in temporal dynamics, our approach uses cross-contrast reference data, which can preserve anatomical structure but inevitably introduces modality-specific features into the target domain, as shown in Fig.~\ref{KCMresult}. To handle these, we introduce a cross-modal feature disentanglement module inspired by the Mamba model~\cite{gu2023mamba} in the image domain. Specifically, we project the modality-mixed images and reference images into a high-dimensional feature space and apply a gating mechanism to extract the modality-specific features. Then, we subtract the reference-specific features from the mixed representation to obtain features aligned with the target modality, thereby suppressing cross-modal interference while preserving features essential for accurate reconstruction. In addition, we design a progressive improvement mechanism to gradually improve disentanglement performance. In summary, the contributions of this paper can be summarized as follows:

\begin{itemize}
\item[1)] We propose a novel method that utilizes fully-sampled reference K-space data to complete undersampled target acquisitions, generating structurally aligned yet modality-mixed inputs while preserving essential anatomical information.
\item[2)] We develop an innovative network architecture leveraging Mamba models to effectively separate and remove reference-specific features from mixed representations, addressing the critical challenge of information contamination in multi-contrast fusion.
\item[3)] We introduce a progressive refinement strategy to cyclically improve the reconstruction quality through repeated feature disentanglement, achieving superior artifact suppression compared to traditional reconstruction methods.
\end{itemize}

\section{Related Work}
\subsection{Multi-Contrast MRI Reconstruction}
Single-contrast methods~\cite{li2021high,li2023multi,yang2023fast,fang2024hfgn} are limited by the amount of available information within a single contrast, which may lead to ambiguity in anatomical structures and poor preservation of fine details. To address these limitations, multi-contrast MRI reconstruction has emerged as a promising strategy by exploiting the complementary information across different contrasts (e.g., T1, T2, PD). Early works~\cite{bilgic2011multi,huang2012fast} have demonstrated improved performance over individual reconstruction by modeling structural consistency across modalities, typically through group sparsity, joint TV regularization, or shared gradient priors.

Some deep learning methods~\cite{xiang2018ultra,zhou20233d,chen2024fefa} use concatenated contrasts as input and fuse the features through neural networks. Dual-path approaches~\cite{falvo2021wide,li2023multi,sun2024multi} employ separate branches for the reference and target modalities, allowing the fusion module to combine multi-scale features extracted from each modality.
In addition, model-driven learning methods~\cite{lei2023decomposition,chen2024multi,lei2024joint,gan2025apg,ding2025enhanced} embed mathematical priors into the network design by unrolling iterative optimization steps guided by predefined MRI objectives, thus emulating traditional reconstruction pipelines within a learnable framework.

However, these methods place their focus on fusing complementary information, without noticing that introducing reference modality will introduce irrelevant features. In contrast, converting multi-modal reconstruction into a disentanglement problem may achieve better results.

\subsection{Dual-Domain MRI Reconstruction}
MRI essentially operates in the frequency domain, where aliasing artifacts are introduced by undersampling of K-space. While image-domain methods treat reconstruction as a visual task, they often discard valuable frequency-domain priors. On the other hand, methods that solely operate in K-space ~\cite{arvinte2021deep,wang2024progressive} often rely on specific sampling patterns, limiting their adaptability and generalization across diverse acquisition settings.

To address the limitations of single-domain strategies, an increasing number of studies have investigated dual-domain frameworks that jointly leverage spatial and frequency information~\cite{eo2018kiki,ran2020md,wang2024ddc,jiang2024memory}. KIKI-net~\cite{eo2018kiki} introduces a hybrid architecture where data consistency (DC) layers are inserted between K-space and image-domain networks, enabling cross-domain feature refinement. Wang et al.~\cite{wang2024ddc} pointed out that sharing the same architecture across domains may hinder domain-specific learning, and proposed customized subnetworks to better extract features from each domain. Ran et al.~\cite{ran2020md} observed that many dual-domain methods process K-space and image-domain data sequentially without explicit interaction. To overcome this, they proposed parallel branches that handle both domains simultaneously with cross-domain communication. Jiang et al.~\cite{jiang2024memory} introduced a contextual memory mechanism that facilitates iterative information exchange, allowing the model to utilize reconstruction knowledge from previous stages to enhance optimization.

However, due to the distinct statistical properties of K-space data, deep learning models designed for visual tasks often fail to generalize well in the frequency domain. Therefore, using the reference modality to complete the target K-space and using deep networks to decouple and reconstruct features in the image domain is a potential solution.

\subsection{Mamba-based MRI Reconstruction}
The Mamba model~\cite{gu2023mamba}, based on a selective state-space mechanism, has recently shown strong potential in both language and vision tasks due to its efficient sequence modeling and linear time complexity. V-Mamba~\cite{liu2024vmamba} extended this idea to vision by introducing a two-dimensional selective scan (SS2D), enabling effective spatial context modeling. Compared to Transformers and CNNs, Mamba offers improved long-range dependency modeling with significantly lower computational complexity and memory usage, making it particularly suitable for high-resolution medical imaging tasks such as MRI reconstruction.

Based on these advances, recent works have begun to explore Mamba for MRI reconstruction~\cite{huang2025enhancing,huang2024mambamir,korkmaz2025mambarecon,ji2024deform,meng2025dm}. Korkmaz et al.~\cite{korkmaz2025mambarecon} proposed a lightweight framework that alternates between Mamba and data consistency blocks. Ji et al.~\cite{ji2024deform} proposed a Deform-Mamba encoder featuring two distinct branches: a modulated deformable block and a vision-oriented Mamba module for feature extraction. Meng et al.~\cite{meng2025dm} introduced a circular scanning approach to better model and retain the structure of the K-space data. Inspired by these works, we aim to integrate the SS2D module into our modality disentanglement module, enabling long-range feature modeling and effective suppression of reference interference in the image domain.

\section{Method}
\subsection{Preliminaries}
\subsubsection{MRI Reconstruction}
MRI reconstruction aims to recover an image $\mathbf{x} \in \mathbb{C}^n$ from its frequency-domain measurements. The fully-sampled K-space $\mathbf{y}^{fs} \in \mathbb{C}^{n}$ is related to the ground-truth image through the following forward model~\cite{lyu2025fast}:
\begin{equation}
    \mathbf{y}^{f}=\mathcal{F} \mathbf{x}+\eta,
\label{equation1}
\end{equation}
where $n = n_{1} \times n_{2}$ indicates the resolution in K-space, $\eta$  represents complex Gaussian noise with distribution $\mathcal{CN}(0,\Sigma)$, $\mathcal{F}$ denotes the two-dimensional Fourier transform.

In the multi-coil setting, $\mathbf{y} = (y_1, y_2, \ldots, y_{n_c})$ denote the full set of undersampled measurements across $n_c$ coils. The forward model can be expressed as:
\begin{equation}
    \mathbf{y}=\mathcal{M} \odot \mathcal{F} \mathbf{S} \mathbf{x}+\eta = \mathbf{A} \mathbf{x}+\eta,
\end{equation}
where $\mathbf{A} = \mathcal{M} \odot \mathcal{F} \mathbf{S}$ is the linear forward operator, $\mathbf{S} = (\mathbf{S}_1, \mathbf{S}_2, \ldots, \mathbf{S}{n_c})$ denotes the coil sensitivity maps, and $\mathcal{M} \odot$ represents an element-wise undersampling mask.

Deep learning-based reconstruction aims to estimate a high-quality image $\hat{\mathbf{x}}$ by learning a mapping from undersampled K-space data $\mathbf{y}$, typically using the following formulation:
\begin{equation}
    \hat{\mathbf{x}}=f_{NET}(\mathrm{IFFT}(\mathbf{y});\theta),
\end{equation}
where $\theta$ denotes parameters of the network and $\mathrm{IFFT}(\mathbf{y})$ represents the zero-filled initial reconstruction obtained by applying the inverse Fourier transform to the undersampled K-space data.

In this work, we focus on the single-coil simulation setting, where the sensitivity maps are not explicitly estimated or used. However, it can be extended to multi-coil reconstruction by incorporating a sensitivity map estimation module or adapting the input pipeline to handle coil-wise data.

\subsubsection{The Mamba Model and SS2D}

Structured State Space Models (SSMs) offer a systematic approach to capturing long-range dependencies through linear dynamics within a latent space. The continuous-time SSM is given by the following set of ordinary differential equations:

\begin{equation}
\left\{
\begin{aligned}
\frac{ d h(t)}{dt} &= A h(t) + B x(t),\\
\quad y(t) &= C h(t) + D x(t),
\end{aligned}
\right.
\label{eqSS2D1}
\end{equation}
where $x(t)$ is the input sequence, $h(t)$ is the latent state, $y(t)$ is the output. $A \in \mathbb{R}^{N \times N}$, $B \in \mathbb{R}^{N \times 1}$, $C \in \mathbb{R}^{1 \times N}$, and $D \in \mathbb{R}$ are learnable parameters. Meanwhile, the term $D x(t)$ serves as a skip connection, enhancing gradient flow and representation capacity.

To incorporate the system into deep networks, zero-order hold (ZOH) discretization is applied with step size $\Delta$. The resulting recurrence formulation is given by:
\begin{equation}
\left\{
\begin{aligned}
&\quad h_k = A h_{k-1} + B x_k,\\
&\quad y_k = C h_k + D x_k,
\end{aligned}
\right.
\label{eqSS2D2}
\end{equation}
where $A = \exp(\Delta A)$ and $B = (\Delta A)^{-1} (\exp(\Delta A) - I) \cdot \Delta B$ are the discretized parameters. This formulation can be further expressed in a convolutional form:
\begin{equation}
\left\{
\begin{aligned}
\quad y &= x \ast K,\\
\quad K &= \left[CB, CAB, \ldots, CA^{L-1}B\right],
\end{aligned}
\right.
\label{eqSS2D3}
\end{equation}
where $L$ is the sequence length and $K \in \mathbb{R}^{L}$ is the SSM convolution kernel used to encode the temporal dynamics of the system.

Base on this foundation, the Mamba model~\cite{gu2023mamba} introduces the Selective State Space (S6) mechanism, which dynamically generates the parameters $B$, $C$, and $\Delta$ in a data-dependent manner. This selective parameterization enables the model to modulate information flow based on the input, enabling efficient and scalable modeling of long-range dependencies with linear time complexity.

To extend SSMs to visual tasks, V-Mamba~\cite{liu2024vmamba} proposes the Selective Scan 2D (SS2D) mechanism, which adapts the selective scanning process to two-dimensional spatial inputs. Unlike traditional one-dimensional scanning that processes tokens sequentially, SS2D performs directional scanning along four spatial directions, thereby capturing contextual information from multiple geometric perspectives. This design bridges the gap between the sequential nature of state space models and the grid-like structure of visual data.

\begin{figure*}[http]
\centering
\includegraphics[width=1\textwidth]{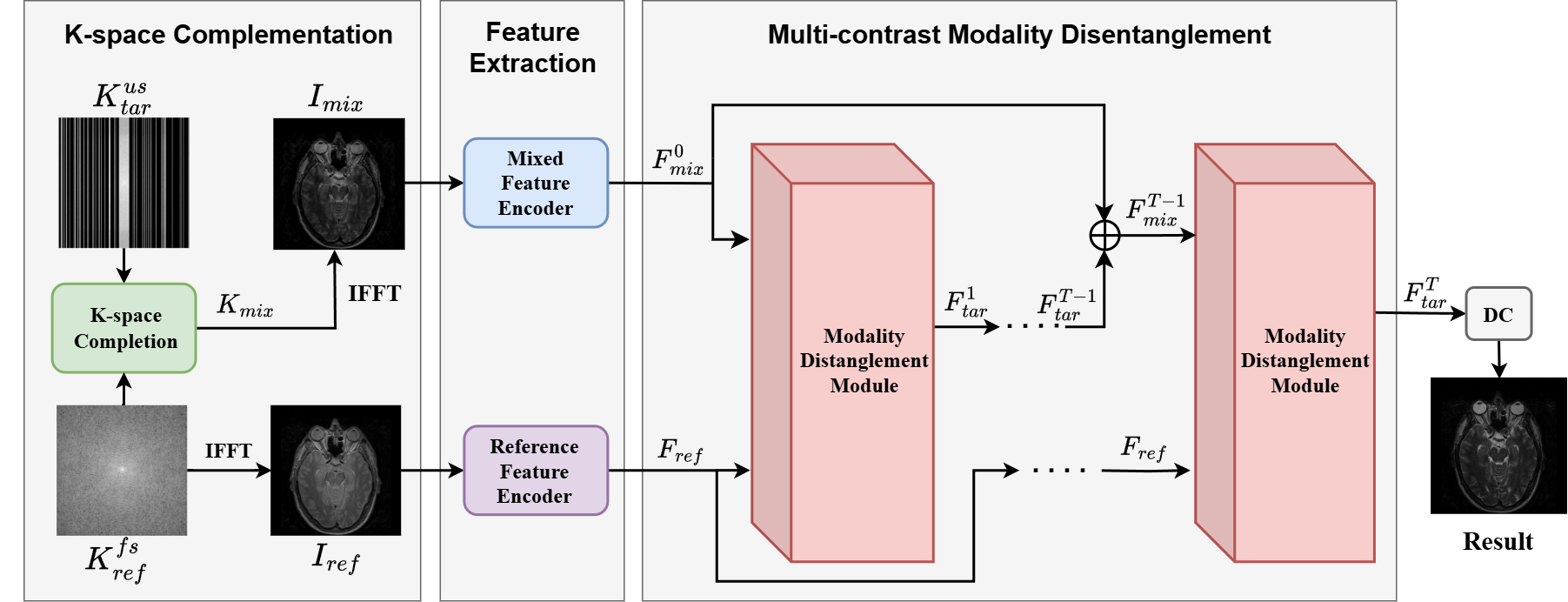} 
\caption{Overview of the proposed MambaMDN framework. The undersampled target K-space data $K_{tar}^{us}$ is complemented by the fully sampled reference data $K_{ref}^{fs}$ to generate a contrast-mixed image $I_{mix}$. Furthermore, we use two encoders to extract high-level features $F_{mix}$ and $F_{ref}$, which are progressively disentangled by Mamba-based modules to reconstruct the target image. Finally, a DC layer is employed to ensure data consistency to get the final result.}
\label{overall_architecture} 
\end{figure*}

\subsection{MambaMDN}
The proposed MambaMDN architecture, illustrated in Fig.~\ref{overall_architecture}, comprises three key stages: (1) K-space complementation, (2) feature extraction, and (3) multi-contrast modality disentanglement. The framework accepts two inputs: undersampled target K-space data and fully-sampled reference K-space data. Specifically, the K-space Complementation Module (KCM) first processes these inputs to generate contrast-mixed K-space data, which is subsequently transformed to the image domain via inverse fast Fourier transform (IFFT). During feature extraction, dedicated modality-specific encoders extract discriminative features from both the mixed representation and reference modality. Finally, the modality disentanglement network, which is based on the Mamba model, leverages the reference features to identify and suppress reference-specific components in the mixed representation, enabling accurate reconstruction of the target modality.

\subsubsection{K-space Complementation Module}
The zero-filled target image is commonly used as the input in many existing reconstruction methods. However, the aliasing artifacts and structural blurring in the image domain severely degrade fine anatomical detail recovery and increase the learning difficulty of the reconstruction network.

\begin{figure}[t]
\centering
\includegraphics[width=0.48\textwidth]{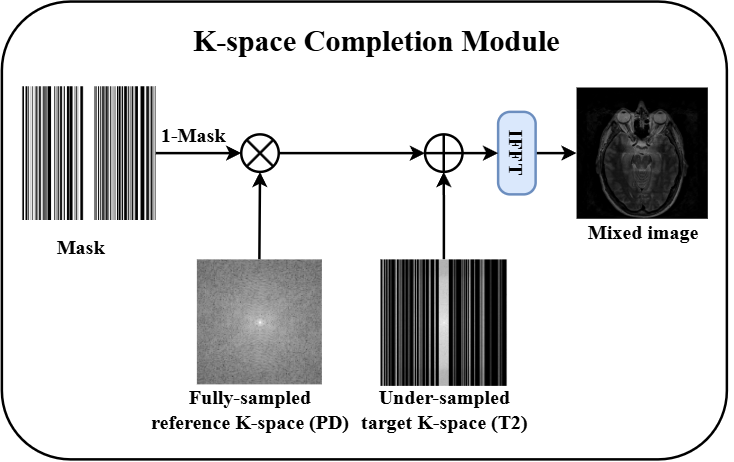} 
\caption{The proposed KCM module. The fully-sampled reference (PD) image is transformed to the K-space domain and used to fill the missing regions of the undersampled target (T2) K-space, yielding an artifact-free but modality-mixed image for the reconstruction network.} 
\label{KCM} 
\end{figure}

To address this limitation, our K-space Complementation Module (KCM) is introduced to generate a modality-mixed image by leveraging the available different contrast information. As shown in Fig.~\ref{KCMresult}, the output of the KCM preserves more structural details and reduces aliasing compared to the zero-filled input, providing a stronger prior for the downstream reconstruction network. In this way, we shift the learning focus from traditional aliasing removal to modality disentanglement. The inner operation is shown in Fig.~\ref{KCM}, which can be summarized by:

\begin{equation}
\left\{
\begin{aligned}
&K_{mix} = K_{tar}^{us} + (1 - M) \odot K_{ref}^{fs},\\
&I_{mix} = \mathrm{IFFT}(K_{mix}),
\end{aligned}
\right.
\label{eqKIC}
\end{equation}
where $M$ is the binary sampling mask, $\odot$ is the element-wise multiplication, $K_{tar}^{us}$ is the input undersampled target data, $K_{ref}^{fs}$ is the fully sampled reference data, and $I_{mix}$ is the feature-mixed image transformed from mixed K-space data $K_{mix}$ via IFFT.

By explicitly complementing the missing K-space regions with structurally similar information from a fully-sampled reference contrast, KCM reduces aliasing and preserves fine anatomical details that are otherwise difficult to recover. This initialization not only guides the network away from overfitting to noise under high acceleration factors (e.g., $\times$8), but also improves the stability and convergence of training.

\begin{figure*}[t]
\centering
\includegraphics[width=0.9\textwidth]{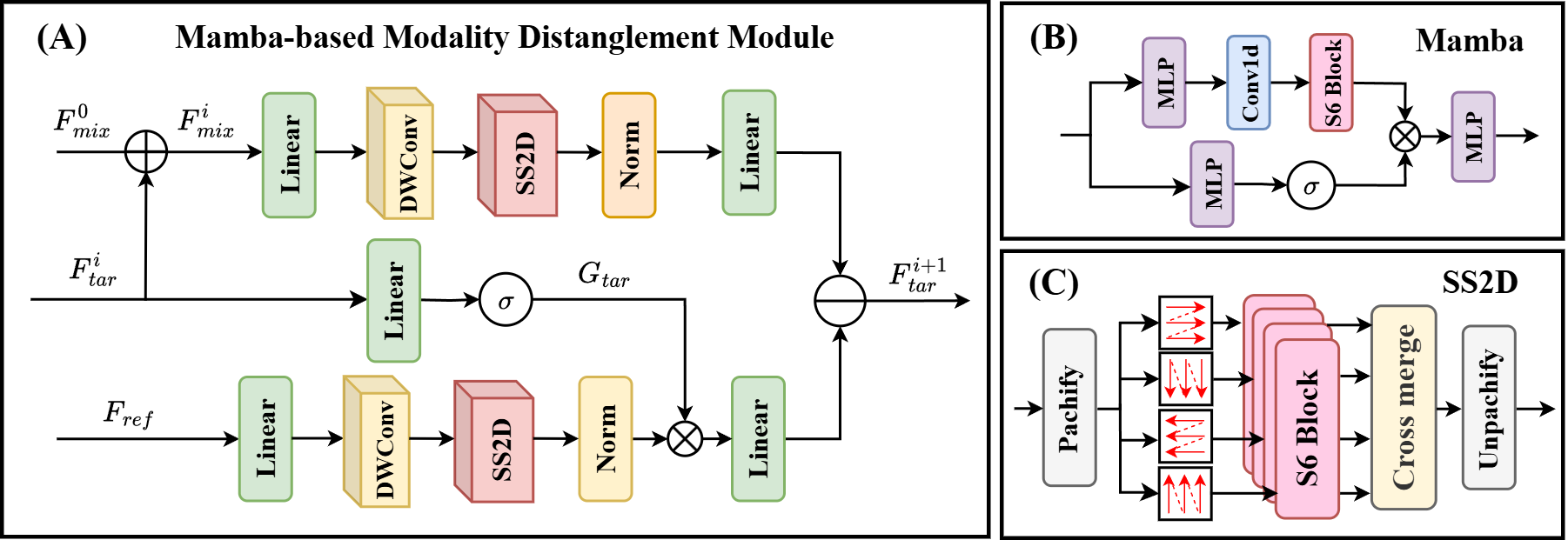} 
\caption{Illustration of the proposed Mamba-based Modality Disentanglement (MMD) module. (A) The overall structure of MMD disentangles modality-specific features from the mixed representation using reference modulation. (B) The original Mamba block. Inspired by it, we designed our MMD module. (C) The SS2D block proposed by Vmamba~\cite{liu2024vmamba} enables the MMD to effectively model spatially long-range features.}

\label{MMD} 
\end{figure*}

\subsubsection{Feature Extraction Module}
To enable more expressive representation learning, we repeat the input images from a single channel to $c$ channels before feeding them into the feature encoders.
Then we employ two encoders: a Mixed Feature Encoder (MFE) to process the feature-mixed image from the KCM module, and a Reference Feature Extractor (RFE) to extract features from the fully sampled reference modality. Although both networks share a U-Net backbone, they serve different purposes:

\begin{itemize}
\item[(1)] The MFE takes the mixed image $I_{mix}$ as input and extracts mixed representations $F_{mix}^{0}$. These features serve as the initial input to the subsequent modality disentanglement network.
\begin{equation}
F_{mix}^{0} = \mathrm{MFE}(I_{mix}).
\label{MFE}
\end{equation}

\item[(2)] The RFE processes the clear reference images $I_{ref}$ to generate a high-level representation $F_{ref}$, which is forwarded to each modality disentanglement block to extract reference-specific features, thereby facilitating more accurate reconstruction of the target modality.
\begin{equation}
F_{ref} = \mathrm{RFE}(I_{ref}).
\label{RFE}
\end{equation}
\end{itemize}

\subsubsection{Mamba-based Modality Disentanglement Module}
On the basis of introducing multi-contrast information to enhance undersampled K-space data, we need to distinguish between target modal features and modal-mixed features. To achieve this, we propose a Mamba-based Modality Disentanglement (MMD) module, as illustrated in Fig.~\ref{MMD} (A). Inspired by the Mamba model~\cite{gu2023mamba} shown in Fig.~\ref{MMD} (B), we extend its original architecture to accept two inputs, and connect the gating branch (with activation $\sigma$) to the reference branch for adaptive modulation. Moreover, to make the module suitable for 2D inputs, we replace the 1D convolution and S6 block with a depthwise convolution and the SS2D block~\cite{liu2024vmamba}, as illustrated in Fig.~\ref{MMD} (C). These adaptations enable the MMD module to progressively filter out reference-specific features, ensuring accurate reconstruction of the target modality. 

Specifically, this module takes the intermediate feature $F_{tar}^{i}$ and the extracted reference feature $F_{ref}$ as input, and outputs a refined, target-specific representation $F_{tar}^{i+1}$. $F_{mix}^{i}$ is obtained by a residual connection between the previous target feature $F_{tar}^{i}$ and the original mixed feature $F_{mix}^{0}$:
\begin{equation}
F_{mix}^{i} = F_{mix}^{0}+F_{tar}^{i}, 0 <i\leq T
\label{eqKIC}
\end{equation}
where $F_{mix}^{i}$ and $F_{tar}^{i}$ are initialized as $F_{mix}^{0}$, which is obtained by the MFE.

Both $F_{mix}^{i}$ and $F_{ref}$ are first passed through the lightweight linear projection. In addition, a depthwise convolution followed by an SS2D block is applied to enhance spatial modeling and long-range contextual interactions. These components ensure structured feature extraction with low computational cost. Meanwhile, a gating mechanism is introduced to adaptively regulate the disentanglement. The modulation gating parameters $G_{tar}^{i}$ is generated via:
\begin{equation}
G_{tar}^{i} = \sigma(\text{Linear}(F_{tar}^{i})),
\end{equation}
where $\sigma(\cdot)$ is an activation function. $G_{tar}^{i}$ can be used to modulate the reference feature, and the gated information is subtracted from the mixed feature path:
\begin{equation}
F_{tar}^{i+1} = \tilde{F}_{mix}^{i} - \text{Linear}(G_{tar} \odot \tilde{F}_{ref}),
\end{equation}
where $\tilde{F}_{mix}^{i}$ and $\tilde{F}_{ref}$ denote the respective features after projection, depthwise convolution, SS2D, and normalization. The symbol $\odot$ denotes element-wise multiplication.

This design enables modality-aware disentanglement by adaptively suppressing irrelevant features from the reference modality while preserving target-specific information. Unlike fusion strategies that risk injecting misleading information from the reference modality, particularly when modality-specific features (e.g., low fluid sensitivity in PD or high contrast of white matter in T1) are inconsistent with the target, our module filters out reference-specific components, ensuring that the final representation is aligned with the target modality. 

Finally, we apply a data consistency (DC) layer after the final MMD block. The DC layer can ensure that the reconstruction conforms to the observed K-space data, learns stably, and preserves fine anatomical details.

\subsubsection{Progressive Refinement Mechanism}
To improve disentanglement performance, we implement a progressive refinement strategy through multi-stage application of our MMD module. At each iteration $i (1  \leq i \leq T)$, the module progressively suppresses reference-specific components from the mixed features using intermediate outputs from the preceding stage. This iterative design enables gradual correction of cross-modal interference, progressively enhancing feature purity and target-modality consistency. Consequently, our MMD module effectively mitigates modality confusion from K-space complementation while improving reconstruction fidelity.

\subsubsection{Loss Function}
Our network is trained in an end-to-end manner, employing the standard pixel-wise L1-loss as the loss function. Specifically, the loss function is applied between the reconstructed target image $\hat{I}_{tar}$ and the fully sampled ground truth data $K_{tar}^{fs}$:
\begin{equation}
\mathcal{L}{L1} = | \hat{I}{tar} - \mathrm{IFFT}(K_{tar}^{fs}) |_1.
\end{equation}
This simple yet effective loss encourages accurate pixel-level recovery and stable convergence.

\section{Experiments}
\subsection{Data Preparation}
\subsubsection{Dataset}
In this work, we use three public multi-contrast datasets to evaluate the effectiveness of the proposed method, including the IXI dataset, the BraTS dataset, and the M4raw dataset.

\textbf{IXI Dataset:} In this dataset, we employ PD modality as the reference and T2 modality as the target. Each modality consists of 576 MRI volumes, which are split into a ratio of 7:1:2 for train, valid, and test sets. The middle 100 slices in each volume are selected for our experiments. Each slice of this dataset has a resolution of 256$\times$256.

\textbf{BraTS:} BraTS~\cite{menze2014multimodal} contains a collection of 285 multi-contrast MR volumes. It was split into a ratio of 7:1:2 for training, validating, and testing. T1 is regarded as the reference modality to direct the reconstruction of the target T2 modality. Similar to the IXI dataset, we only utilize the middle 100 slices for our experiments, and each slice has a resolution of 240$\times$240.

\textbf{M4raw:} M4raw~\cite{lyu2023m4raw} is a multi-contrast and multi-channel MRI dataset containing T1, T2, and FLAIR contrast. In our experiments, we filter out 150 paired T1 and T2 brain volumes for our model evaluation, and each volume has 18 slices with 256$\times$256 resolution. These slices are split into a ratio of 7:1:2 for training, validating, and testing. T1 modality is directed toward the restoration of T2 modality.

\subsubsection{Mask}
In this work, we adopt a variable-density random sampling mask along the phase-encoding direction to simulate undersampled K-space data. Specifically, we use acceleration factors of $\times$4 and $\times$8 to evaluate the model under both moderate and high undersampling scenarios. Each mask retains a fixed low-frequency region in the center to preserve structural information, while the remaining frequencies are randomly sampled with a probability determined by the desired acceleration rate. A fixed random seed is used to ensure reproducibility.

\subsection{Evaluation Metrics and Experimental Setup}
In this work, we quantitatively evaluated our model using three standard metrics: peak signal-to-noise ratio (PSNR), structural similarity index (SSIM), and root mean square error (RMSE). These metrics were computed on image-domain reconstructions obtained through inverse Fourier transform of the processed K-space data. Following convention, superior reconstruction quality is indicated by higher PSNR/SSIM values and lower RMSE values.

Our experimental implementation utilizes the PyTorch framework running on NVIDIA GeForce RTX 3090 GPUs. In the training phase, we employed the Adam optimizer with a learning rate of $1 \times 10^{-4}$ and batch size of 6 for 50 training epochs. For the final model, the channel number $c$ is expanded to 32 after the KCM module, and the number of MMD modules is set to 6. 

\begin{table*}[htbp]
\centering
\scriptsize
\caption{Quantitative comparison on IXI, BraTS, and M4raw datasets under ×4 and ×8 acceleration rates. Best results and second-best results are highlighted and underlined, respectively.}
\renewcommand{\arraystretch}{0.9}
\setlength{\tabcolsep}{2.1mm}
\normalsize
\begin{tabular}{cccccccc}
\toprule
\textbf{Dataset} & \textbf{Methods} & \multicolumn{3}{c}{×4 Acceleration} & \multicolumn{3}{c}{×8 Acceleration} \\
\cmidrule(lr){3-5} \cmidrule(lr){6-8}
& & PSNR (dB) $\uparrow$ & SSIM $\uparrow$ & RMSE $\downarrow$ & PSNR (dB) $\uparrow$ & SSIM $\uparrow$ & RMSE $\downarrow$ \\
\midrule
\multirow{8}{*}{IXI} 
& Zero-Filled (ZF) & 24.60 & 0.5420 & 15.94 & 23.39 & 0.4995 & 18.31 \\
& U-Net \cite{ronneberger2015u} (MICCAI 2015) & 31.95 & 0.9032 & 6.933 & 30.66 & 0.8875 & 7.980 \\
& Restormer \cite{zamir2022restormer} (CVPR 2022) & 38.10 & 0.9693 & 3.417 & 36.46 & 0.9629 & 4.046 \\
& MC-VANet \cite{lei2023decomposition} (ICCV 2023) & \underline{40.77} & \underline{0.9778} & \underline{2.510} & \underline{38.20} & \underline{0.9681} & \underline{3.354} \\
& PanMamba \cite{he2025pan} (IF 2025) & 39.01 & 0.9703 & 3.048 & 37.56 & 0.9650 & 3.602 \\
& MambaRecon \cite{korkmaz2025mambarecon} (WACV 2025) & 39.66 & 0.9708 & 2.838 & 36.76 & 0.9588 & 3.918 \\
& MambaMMR \cite{zou2025mmr} (MedIA 2025) & 38.41 & 0.9636 & 3.259 & 37.10 & 0.9632 & 3.747 \\
& \textbf{MambaMDN (ours)} & \textbf{41.92} & \textbf{0.9806} & \textbf{2.211} & \textbf{39.30} & \textbf{0.9724} & \textbf{2.981} \\
\midrule
\multirow{8}{*}{BraTS}
& Zero-Filled (ZF) & 24.65 & 0.5642 & 15.17 & 24.37 & 0.5630 & 15.66 \\
& U-Net \cite{ronneberger2015u} (MICCAI 2015) & 30.90 & 0.9140 & 7.404 & 29.27 & 0.8905 & 8.944 \\
& Restormer \cite{zamir2022restormer} (CVPR 2022) & 34.21 & 0.9506 & 5.071 & 33.24 & 0.9429 & 5.667 \\
& MC-VANet \cite{lei2023decomposition} (ICCV 2023) & 37.75 & 0.9699 & 3.390 & \underline{35.55} & \underline{0.9573} & 4.364 \\
& PanMamba \cite{he2025pan} (IF 2025) & 36.29 & 0.9659 & 4.005 & 34.84 & 0.9567 & 4.732 \\
& MambaRecon \cite{korkmaz2025mambarecon} (WACV 2025) & 36.84 & 0.9594 & 3.745 & 34.87 & 0.9484 & 4.695 \\
& MambaMMR \cite{zou2025mmr} (MedIA 2025) & \underline{37.92} & \underline{0.9709} & \underline{3.292} & 35.54 & 0.9568 & \underline{4.330} \\
& \textbf{MambaMDN (ours)} & \textbf{38.18} & \textbf{0.9727} & \textbf{3.228} & \textbf{35.84} & \textbf{0.9606} & \textbf{4.214} \\
\midrule
\multirow{8}{*}{M4raw}
& Zero-Filled (ZF) & 24.36 & 0.6014 & 15.52 & 23.14 & 0.5141 & 17.86 \\
& U-Net \cite{ronneberger2015u} (MICCAI 2015) & 28.96 & 0.7572 & 9.150 & 26.46 & 0.6639 & 12.20 \\
& Restormer \cite{zamir2022restormer} (CVPR 2022) & 30.84 & 0.7692 & 7.367 & 29.08 & 0.7219 & 9.021 \\
& MC-VANet \cite{lei2023decomposition} (ICCV 2023) & \underline{31.82} & \underline{0.8161} & \underline{6.576} & 28.70 & 0.7107 & 9.424 \\
& PanMamba \cite{he2025pan} (IF 2025) & 31.58 & 0.8146 & 6.759 & \underline{29.13} & \underline{0.7276} & \underline{8.973} \\
& MambaRecon \cite{korkmaz2025mambarecon} (WACV 2025) & 29.42 & 0.7391 & 8.673 & 27.11 & 0.6380 & 11.30 \\
& MambaMMR \cite{zou2025mmr} (MedIA 2025) & 29.52 & 0.7507 & 8.576 & 28.19 & 0.7035 & 9.994 \\
& \textbf{MambaMDN (ours)} & \textbf{32.01} & \textbf{0.8205} & \textbf{6.435} & \textbf{29.75} & \textbf{0.7310} & \textbf{8.347} \\ 
\bottomrule
\end{tabular}
\label{results}
\end{table*}

\begin{figure*}[t]
\centering
\includegraphics[width=0.975\textwidth]{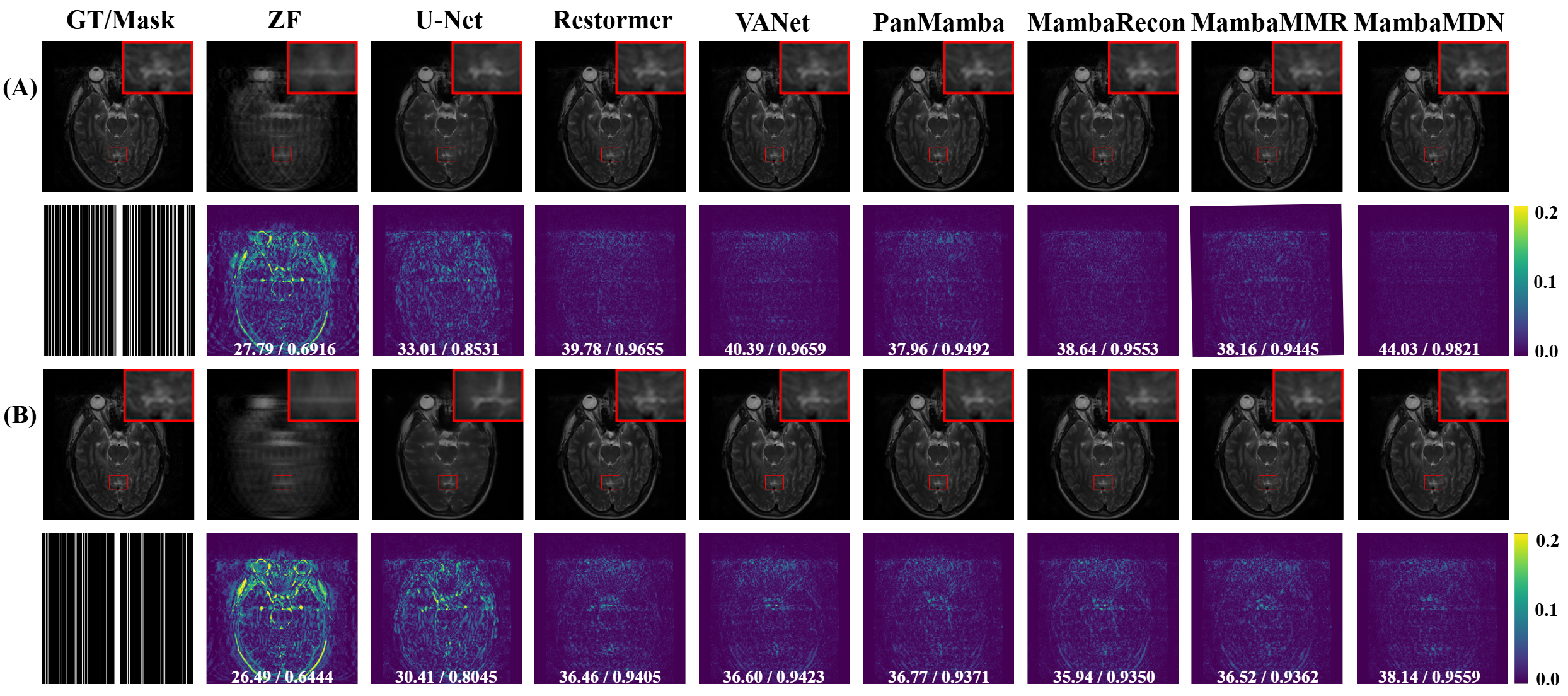} 
\caption{Visual comparison of different methods on the IXI dataset. The first row of each group shows the reconstruction result, while the second row indicates the error map. Group (A) uses a 1D random undersampling mask with $\times$4 acceleration rate, while Group (B) uses a $\times$8 rate mask.}
\label{cI} 
\end{figure*}

\begin{figure*}[t]
\centering
\includegraphics[width=0.975\textwidth]{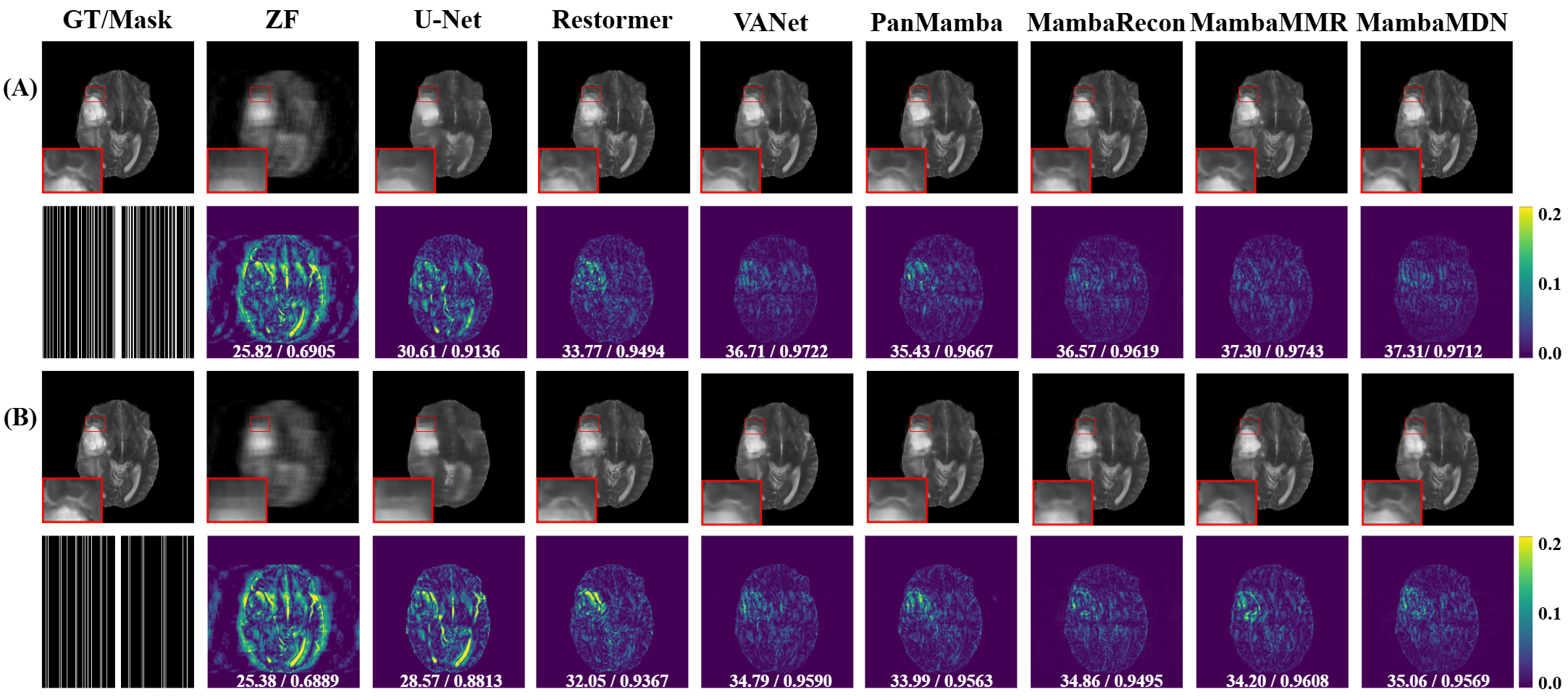} 
\caption{Visual comparison of different methods on the BraTS dataset. The first row of each group shows the reconstruction result, while the second row indicates the error map. Group (A) uses a 1D random undersampling mask with $\times$4 acceleration rate, while Group (B) uses a $\times$8 rate mask.}
\label{cB} 
\end{figure*}

\subsection{Quantitative Comparison}

\begin{figure*}[t]
\centering
\includegraphics[width=1\textwidth]{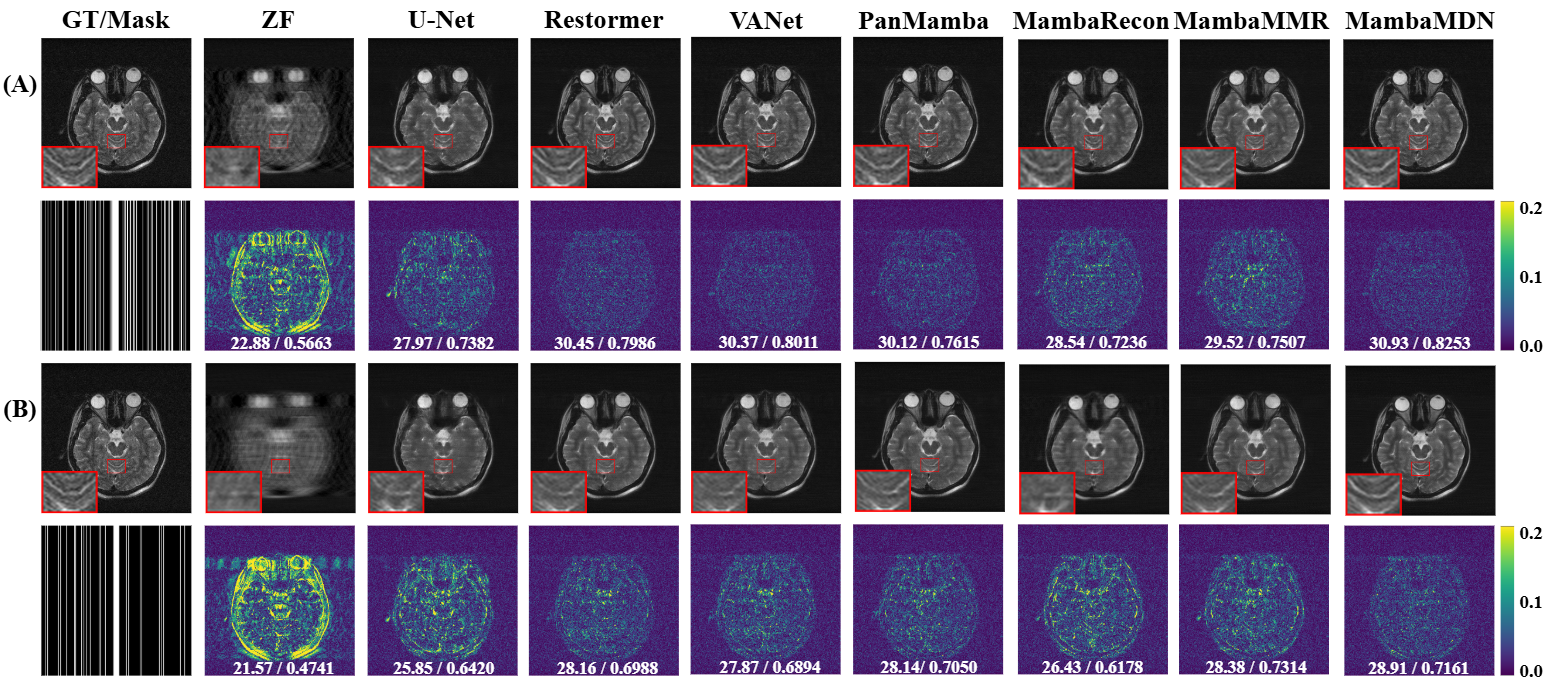} 
\caption{Visual comparison of different methods on the M4raw dataset. The first row of each group shows the reconstruction result, while the second row indicates the error map. Group (A) uses a 1D random undersampling mask with $\times$4 acceleration rate, while Group (B) uses a $\times$8 rate mask.}
\label{cM} 
\end{figure*}

We conducted a comparative evaluation of the proposed MambaMDN against several existing MRI reconstruction methods to demonstrate its effectiveness, including single-contrast baselines such as U-Net~\cite{ronneberger2015u}, Restormer~\cite{zamir2022restormer}, and MambaRecon~\cite{korkmaz2025mambarecon}, as well as multi-contrast methods like MC-VANet~\cite{lei2023decomposition}, PanMamba~\cite{he2025pan}, and MambaMMR~\cite{zou2025mmr}. TABLE~\ref{results} reports the quantitative results of these models on IXI, BraTS, and M4raw datasets under $\times$4 and $\times$8 acceleration settings.

As shown in TABLE~\ref{results}, our MambaMDN achieves the best performance across all metrics and datasets. For instance, on the IXI dataset with $\times$4 acceleration, MambaMDN surpasses MC-VANet and MambaRecon by 1.15 dB and 2.26 dB in PSNR, respectively. On the more challenging $\times$8 setting, MambaMDN still maintains a clear margin, outperforming MC-VANet by 1.10 dB. Similarly, on the BraTS and M4raw dataset, our method achieves the highest PSNR and SSIM under both acceleration factors. It is worth noting that although MambaMMR~\cite{zou2025mmr} and MC-VANet~\cite{lei2023decomposition} are also multi-contrast reconstruction methods, their performance lags behind our MambaMDN. This highlights the effectiveness of our modality disentanglement strategy, which allows MambaMDN to better exploit complementary information from the reference modality while preserving target-specific features. 

\subsection{Qualitative Results}
Figs.~\ref{cI}, \ref{cB}, and \ref{cM} present qualitative comparisons of these models on IXI, BraTS, and M4raw datasets, respectively. Group (A) corresponds to 1D random undersampling with $\times$4 acceleration rate, while Group (B) corresponds to $\times$8 setting. In addition, the first row of each group shows the reconstruction result, while the second row indicates the error map. Obviously, our MambaMDN achieves superior preservation of structural details and significantly reduces reconstruction errors compared to other methods. The proposed framework consistently maintains high-fidelity texture representation and precise anatomical boundary delineation across varying undersampling rates, exhibiting remarkable robustness and versatile adaptation to different acquisition scenarios. 

These improvements are not only beneficial in a technical sense but also particularly meaningful in clinical contexts, where accurate visualization of anatomical structures is essential for reliable interpretation. This is especially evident in the IXI dataset, which comprises healthy brain scans with clear and consistent structural features. Our method performs notably well in this setting, highlighting its strong capacity to preserve fine-grained anatomical details, which is important for clinical scenarios that rely on high-resolution imaging of normal tissue morphology.

\subsection{Model Complexity}
As shown in TABLE~\ref{TrainingEfficiency}, we compared the FLOPs, memory consumption, and reconstruction performance of multiple models based on Mamba and Transformer on the IXI dataset (batch size=4). Notably, Mamba-based methods (PanMamba, MambaRecon, MambaMMR, and our MambaMDN) exhibit superior computational efficiency. Our MambaMDN achieves the best reconstruction result with moderate resource requirements, demonstrating its effectiveness in joint spatial-modality feature modeling. Compared to Transformer-based Restormer and model-driven MC-VANet, MambaMDN reduces computational costs by 30.7\% and 74.1\% respectively while delivering superior reconstruction fidelity. Therefore, we can draw a conclusion that our MambaMDN achieves the best balance between model complexity and performance.

\begin{table}[t]
\centering
\renewcommand{\arraystretch}{1.1}
\setlength{\tabcolsep}{2mm}
\small
\caption{Quantitative comparison of model complexity (FLOPs, Memory, and reconstruction quality) on the IXI dataset with batch size = 4.}
\begin{tabular}{lcccc}
\toprule
\textbf{Methods} & \textbf{FLOPs} & \textbf{Memory} & \textbf{PSNR} & \textbf{SSIM}  \\
\midrule
U-Net \cite{ronneberger2015u} & 35.86G & 1.42GB & 31.95 & 0.9032 \\
Restormer \cite{zamir2022restormer} & 161.92G & 26.42GB & 38.10 & 0.9693 \\
PanMamba \cite{he2025pan} & 76.17G & 10.43GB & 39.01 & 0.9703 \\
MambaRecon \cite{korkmaz2025mambarecon} & 30.29G & 5.66GB & 39.66 & 0.9708 \\
MambaMMR \cite{zou2025mmr} & 99.41G & 11.98GB & 38.41 & 0.9636 \\
MC-VANet \cite{lei2023decomposition} & 432.62G & 12.55GB & 40.77 & 0.9778 \\
MambaMDN (ours) & 112.15G & 7.86GB & \textbf{41.92} & \textbf{0.9806} \\
\bottomrule
\end{tabular}
\label{TrainingEfficiency}
\end{table}

\begin{table}[t]
\centering
\renewcommand{\arraystretch}{1.1}
\setlength{\tabcolsep}{3.5mm}
\small
\caption{Comparison of different modality disentanglement strategies on the IXI dataset. The best results are highlighted.}
\begin{tabular}{lccc}
\toprule
\textbf{Strategy} & \textbf{PSNR$\uparrow$} & \textbf{SSIM$\uparrow$} & \textbf{RMSE$\downarrow$} \\
\midrule
$F_{mix} - F_{ref}$ (ours) & \textbf{41.92} & \textbf{0.9806} & \textbf{2.211} \\
$F_{tar}^{us} - F_{ref}$ & 41.48 & 0.9789 & 2.327 \\
$F_{mix} - F_{tar}^{us}$ & 41.13 & 0.9762 & 2.411 \\
$F_{mix} - 0$ & 41.33 & 0.9649 & 2.355 \\
\bottomrule
\end{tabular}
\label{disentangle}
\end{table}

\section{Ablation Study}
\subsection{Effectiveness of Modality Disentanglement Strategy}

\begin{figure}[t]
\centering
\includegraphics[width=0.48\textwidth]{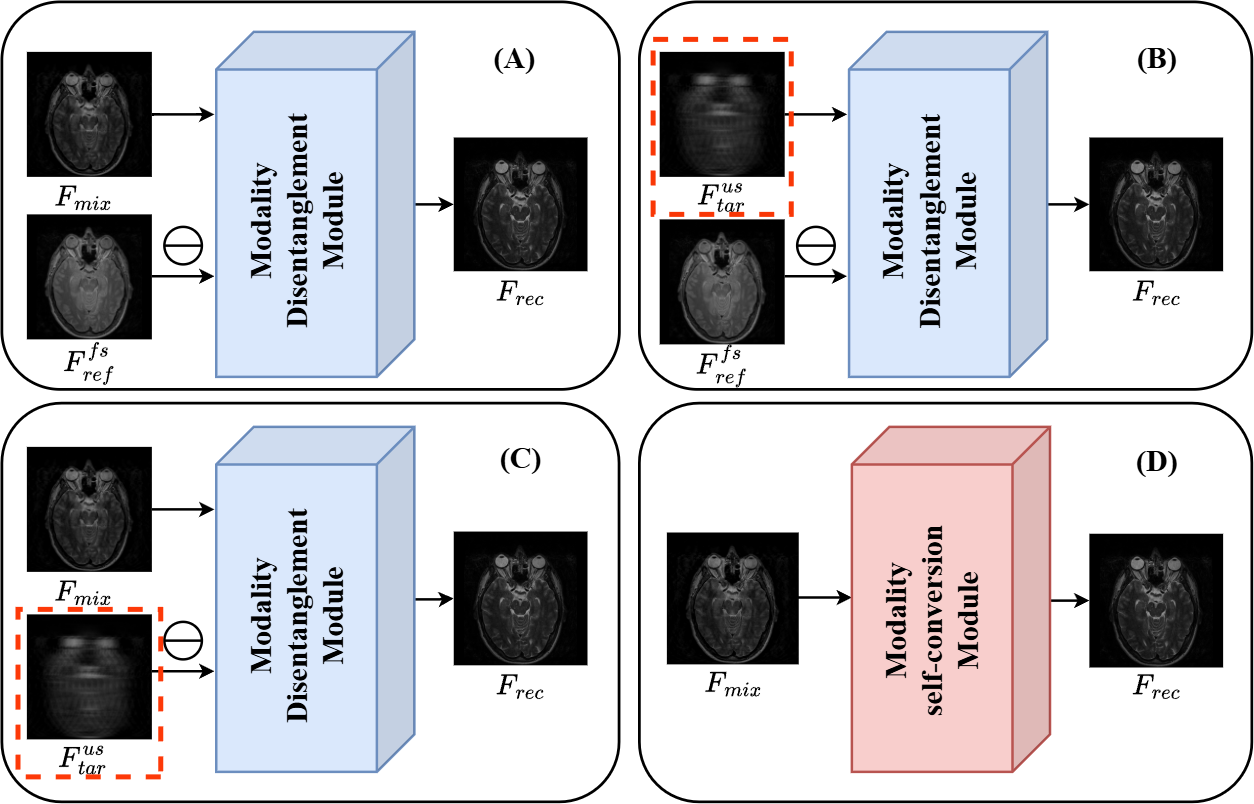} 
\caption{Ablation study on disentanglement designs. (A) Our full model subtracts the reference features from the mixed modality. (B) Replacing the KCM-generated input with the undersampled target. (C) Replacing the subtraction input with the undersampled target leads to inaccurate disentanglement. (D) Removing the reference entirely forces the model to perform modality self-conversion.} 
\label{ablation} 
\end{figure}

\begin{figure}[t]
\centering
\includegraphics[width=0.48\textwidth]{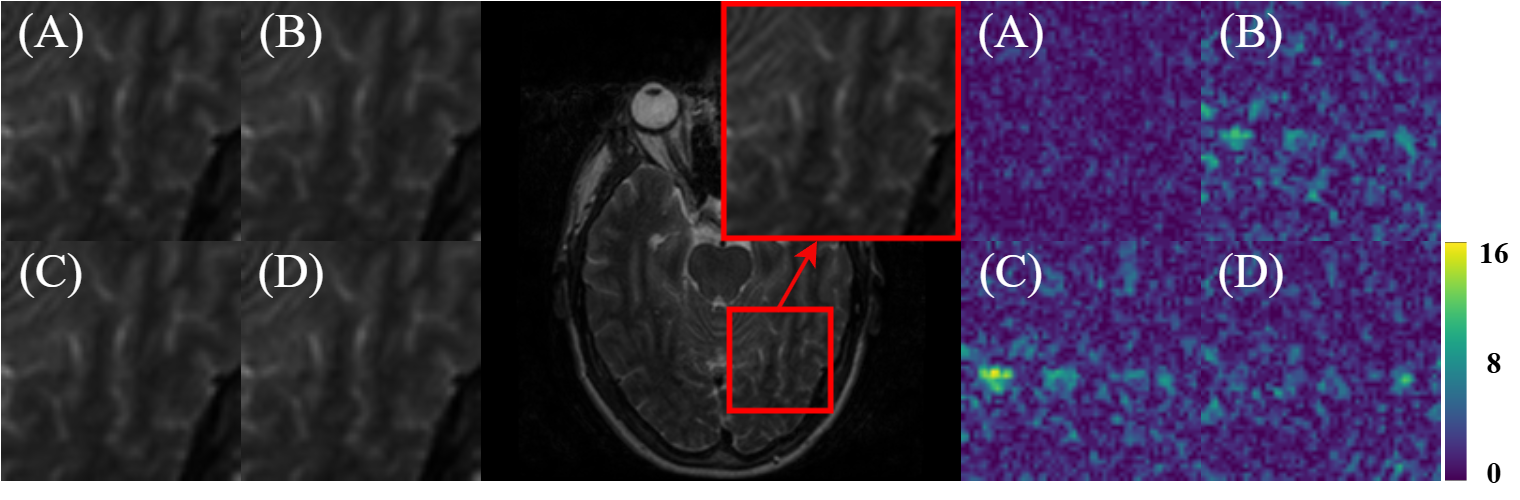} 
\caption{Result of different modality disentanglement strategies. Our method achieved the best reconstruction result.} 
\label{ablationvisible} 
\end{figure}

To evaluate the effectiveness of our proposed modality disentanglement strategy, we conduct ablation studies by comparing different disentanglement designs, as illustrated in Fig.~\ref{ablation}. The quantitative results on the IXI dataset are summarized in TABLE~\ref{disentangle}, while Fig.~\ref{ablationvisible} presents qualitative comparisons of the reconstructed images.

$F_{mix} - F_{ref}$ (ours): Our full method separates the target-relevant features by removing reference-specific information from the mixed representation, using fully sampled reference data. This produces the best results as it can accurately suppress irrelevant structures and better preserve target-specific content.

$F_{tar}^{us} - F_{ref}$: In this setting, we replace the KCM-generated input with the undersampled target modality directly. This lead to a 0.44 drop in PSNR and a 0.017 decrease in SSIM. Since the gating mechanism is able to preserve target features and suppress reference interference, the disentanglement module still works even when starting from an aliased target. However, the presence of noise and aliasing in the undersampled target can hinder the gating's ability to distinguish between useful structures and artifacts, thus reducing the overall quality of the disentanglement.
    
$F_{mix} - F_{tar}^{us}$: This design subtracts the undersampled target features from the mixed representation. Theoretically, this operation will get reference-specific features, which is contradictory to our goal of reconstructing the target modality. As a result, the reconstruction performance drops the most, with a decrease of 0.79 dB in PSNR and 0.044 in SSIM.

(D) $F_{mix} - 0$: In this case, we modified the MMD module into a single-contrast version, forcing the model to perform modality self-conversion without any reference-specific guidance. Thanks to the de-aliased and structurally clean input provided by the KCM module, the model maintained reasonable pixel-level accuracy, evidenced by only a modest PSNR drop of 0.59 dB. However, the lack of reference guidance introduced noticeable cross-modal interference, significantly compromising structural fidelity and leading to a substantial SSIM reduction of 0.0157.

\subsection{Impact of MMD Module Quantity on Performance}
To evaluate the impact of the MMD module count on reconstruction performance, we conducted ablation experiments on the IXI test set. As summarized in TABLE~\ref{MMDnumber}, increasing the number of MMD modules from 4 to 6 leads to progressive improvements in PSNR and SSIM alongside reduced RMSE, demonstrating enhanced image quality and more effective modality disentanglement. However, increasing to 7 modules causes performance degradation, likely due to feature over-subtraction.  Based on these findings, we use 6 MMD modules in the final model to maximize reconstruction quality. Notably, the Mamba-based architecture’s lightweight design ensures this configuration maintains high computational efficiency without introducing substantial overhead.

\subsection{Effectiveness of Iterative Refinement Strategy}

To investigate the impact of the iterative refinement strategy on model performance, we designed an ablation experiment to compare the results with and without this strategy. Specifically, we conduct an ablation study where 6 MMD modules are employed without progressive refinement strategy. We stack 6 MMD modules in a parallel way, and they independently process the same input pair $(F_{mix}^0, F_{ref})$, then the outputs are averaged to get the final target result. As shown in Table 2, the performance of both 'single block' or '6 parallel blocks' has decreased. In contrast, the model using our proposed iterative refinement strategy achieved the best results. This fully validates the effectiveness of the strategy.

\begin{table}[t]
\centering
\renewcommand{\arraystretch}{1.1}
\setlength{\tabcolsep}{5.5mm}
\small
\caption{Impact of MMD Module Quantity on Performance.}
\begin{tabular}{lccc}
\toprule
\textbf{Number} & \textbf{PSNR$\uparrow$} & \textbf{SSIM$\uparrow$} & \textbf{RMSE$\downarrow$} \\
\midrule
4 blocks   & 41.51 & 0.9789 & 2.312 \\
5 blocks   & 41.69 & 0.9797 & 2.266 \\
6 blocks   & \textbf{41.92} & \textbf{0.9806} & \textbf{2.211} \\
7 blocks   & 41.83 & 0.9802 & 2.231 \\
\bottomrule
\end{tabular}
\label{MMDnumber}
\end{table}

\begin{table}[t]
\centering
\renewcommand{\arraystretch}{1.1}
\setlength{\tabcolsep}{3.2mm}
\small
\caption{Results of the model with and without iterative refinement strategy on the IXI test set.}
\begin{tabular}{lccc}
\toprule
\textbf{Parallel or Iterative} & \textbf{PSNR$\uparrow$} & \textbf{SSIM$\uparrow$} & \textbf{RMSE$\downarrow$} \\
\midrule
    6 iterative blocks (ours)  & \textbf{41.92} & \textbf{0.9806} & \textbf{2.211} \\
    6 parallel blocks   & 39.57 & 0.9713 & 2.860 \\
    single block        & 38.97 & 0.9683 & 3.063  \\

\bottomrule
\end{tabular}
\label{ablationiterative}
\end{table}

\subsection{Effectiveness of K-space Complementation Module}

To evaluate the effectiveness of the proposed K-space Complementation Module (KCM), we removed the KCM from our MMD framework. In this setting, the model takes the undersampled target image $I_{tar}^{us}$ and the fully sampled reference image $I_{ref}^{fs}$ as input. We modify our disentanglement module to perform a fusion strategy, replacing the subtraction operation with feature addition. As shown in the last row of TABLE~\ref{ablationKCM}, removing the KCM led to a noticeable drop in performance, 0.17 dB in PSNR and 0.0012 in SSIM, indicating that the input from K-space complementation provides more informative guidance than direct fusion.

\begin{figure}[t]
\centering
\includegraphics[width=0.48\textwidth]{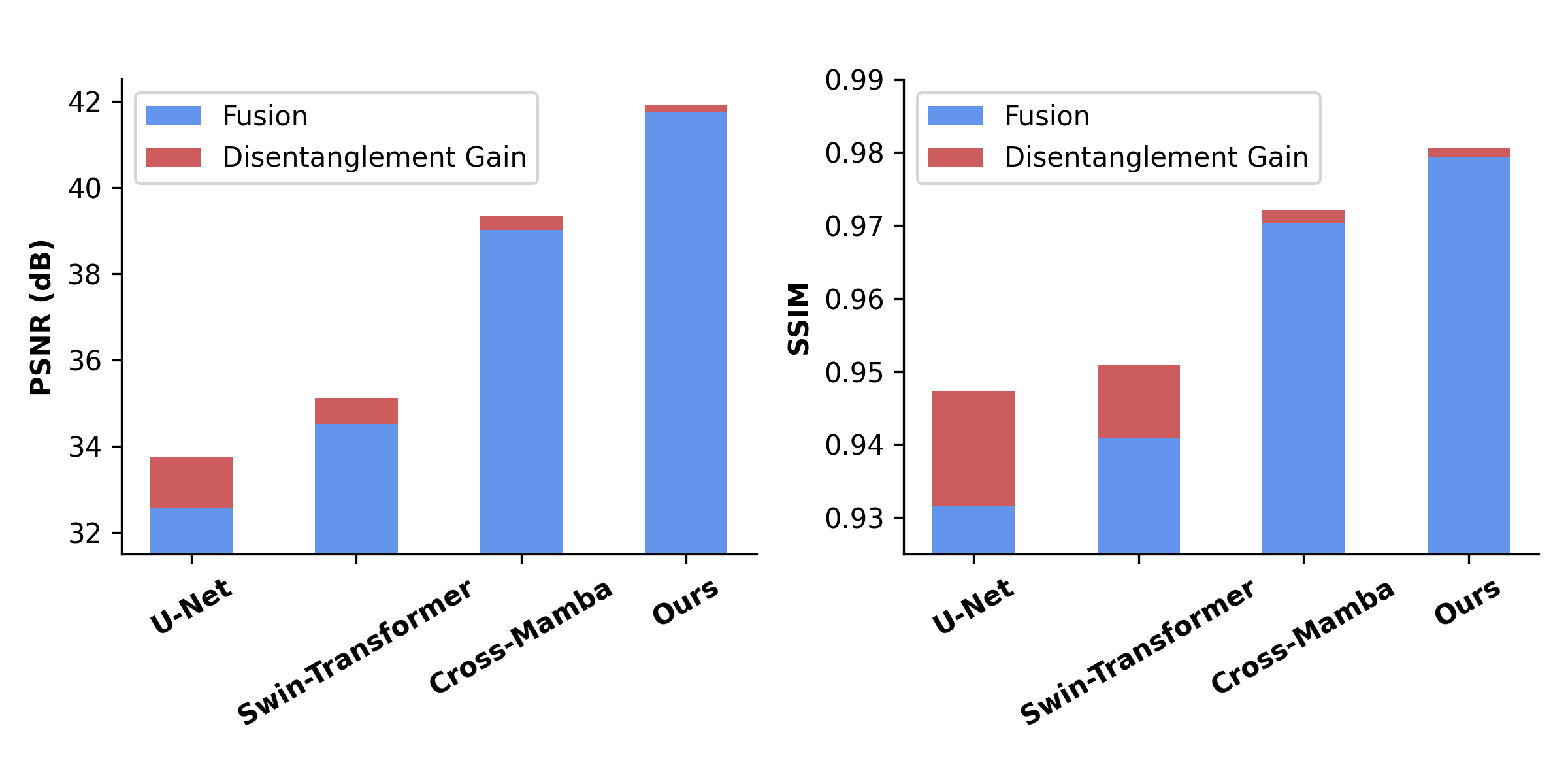} 
\caption{Comparison of fusion and disentanglement strategies across different backbone architectures.} 
\label{FvsD} 
\end{figure}

Meanwhile, we evaluated the effectiveness of KCM across multiple backbone architectures, including U-Net~\cite{ronneberger2015u}, Swin Transformer~\cite{liu2021swin}, and Cross-Mamba~\cite{he2025pan}. For each backbone, we designed two variants: one that simply fuses (Fusion) and another that leverages the proposed KCM to produce a modality-mixed image, followed by a modality disentanglement strategy (Disentanglement). As shown in Fig.~\ref{FvsD} and TABLE~\ref{ablationKCM}, among all architectures, the KCM-based disentanglement method consistently outperforms simple fusion. By directly combining the low-quality target K-space and the high-quality reference K-space using the undersampling mask, KCM reconstructs a cleaner and artifact-suppressed mixed image, which serves as a better input for downstream feature interaction. Notably, the improvements are more prominent in lighter architectures (e.g., U-Net and Swin Transformer), highlighting the generality and effectiveness of KCM as an enhancement module.

\begin{table}[t]
\centering
\renewcommand{\arraystretch}{1.1}
\setlength{\tabcolsep}{1mm}
\footnotesize
\caption{Comparison between different networks in the feature interaction stage.}
\begin{tabular}{lcccccc}
\toprule
\multirow{2}{*}{\textbf{Methods}} & \multicolumn{3}{c}{\textbf{Disentanglement}} & \multicolumn{3}{c}{\textbf{Fusion}} \\
& \textbf{PSNR $\uparrow$} & \textbf{SSIM $\uparrow$} & \textbf{RMSE $\downarrow$} & \textbf{PSNR $\uparrow$} & \textbf{SSIM $\uparrow$} & \textbf{RMSE $\downarrow$} \\
\midrule
U-Net  \cite{ronneberger2015u}     &   33.75  &  0.9473  &  5.490  & 32.58 & 0.9316 & 6.260 \\
SwinT\cite{liu2021swin} & 35.12 & 0.9509&  4.720 & 34.51 & 0.9409 &  5.083 \\
Cross-M\cite{he2025pan}  & 39.34 & 0.9721  & 2.939 & 39.01 & 0.9703 & 3.048   \\
Ours    & \textbf{41.92} & \textbf{0.9806} & \textbf{2.211} & \textbf{41.75} &  \textbf{0.9794} & \textbf{2.255} \\
\bottomrule
\end{tabular}
\label{ablationKCM}
\end{table}

\section{Discussion}
In this study, we present MambaMDN, an innovative reconstruction paradigm that transforms multi-modal MRI from conventional feature fusion to anatomically-aware modality disentanglement. This approach addresses two problems: (1) our k-space complementation strategy helps preserve native tissue contrast during acquisition, reducing aliasing artifacts in the reconstructed images, and (2) the Mamba-based disentanglement module reduces interference from reference anatomy, which may improve feature specificity and aid lesion visualization.

Regarding potential clinical use, MambaMDN demonstrates the ability to reconstruct target contrasts (e.g., T2-weighted) from substantially undersampled acquisitions when aided by reference scans (e.g., T1-weighted). Meanwhile, the computational efficiency of the method makes it promising for clinical deployment. Compared to existing solutions requiring GPU clusters, MambaMDN offers a favorable balance between reconstruction performance and resource consumption, which is especially valuable in time-sensitive or resource-limited settings such as rural clinics, emergency treatment, and low-field MRI systems.

Although our MambaMDN demonstrates strong reconstruction performance, several limitations remain. First, residual cross-modal contamination still persists in the critical anatomical region, and conventional pixel-wise metrics inadequately capture diagnostic fidelity for pathological tissues, suggesting the need for lesion-aware optimization criteria that prioritize clinically relevant features like tumor margins or edema delineation. Second, downstream tasks such as lesion visualization, segmentation, and lesion-level analysis, were not conducted in this work, and the framework’s current validation lacks assessment of real-world variability across scanner vendors, field strengths (1.5T vs. 3T), and institutional protocols. Future work will integrate uncertainty quantification for lesion boundaries and multi-center studies to address these issues.

\section{Conclusion}
In this paper, we proposed MambaMDN, a novel dual-domain MRI reconstruction framework that synergizes K-space complementation with modality disentanglement for multi-contrast MRI reconstruction. Unlike existing methods that rely on fusion rules, our method employs a two-stage strategy: First, we leverage fully-sampled reference K-space data to complete the undersampled target K-space, generating a de-aliased initialization that preserves anatomical structures. Second, a Mamba-based modality disentanglement module progressively eliminates reference-specific features through adaptive filtering, ensuring target-specific reconstruction fidelity. Extensive experiments on public MRI datasets demonstrate that our method consistently outperforms state-of-the-art techniques in terms of both quantitative metrics and visual quality. In summary, our work provides a new perspective on integrating cross-modal priors for high-fidelity MRI reconstruction and lays a foundation for future extensions to clinically-oriented applications.

\bibliographystyle{unsrt}
\bibliography{main.bbl}
\end{document}